%% file: expertise.tex
\pdfoutput=1

\documentclass[11pt]{article}

\usepackage[final]{style/acl}

\usepackage{times}
\usepackage{latexsym}

\usepackage[T1]{fontenc}

\usepackage[utf8]{inputenc}

\usepackage{microtype}

\usepackage{inconsolata}

\usepackage{svg}
\usepackage{outlines}
\usepackage{graphicx}
\usepackage{siunitx}
\usepackage{amsfonts}
\usepackage{amsmath}
\usepackage{booktabs}
\usepackage{multirow}

\usepackage{graphicx}
\usepackage{algorithm}
\usepackage{enumitem}
\usepackage{tablefootnote}
\usepackage{algpseudocode}
\algrenewcommand\algorithmicrequire{\textbf{Input:}}
\algrenewcommand\algorithmicensure{\textbf{Output:}}

\usepackage{caption}
\usepackage{bbold}
\usepackage{cleveref}
\usepackage{subcaption}
\usepackage{soul, color}
\crefname{section}{§}{§§}
\input{style/preamble.tex}

\title{Speaking the Right Language: The Impact of Expertise \\Alignment in User-AI Interactions}

\newcommand{\instA}{\dagger}
\newcommand{\instB}{\diamondsuit}
\newcommand{\authorsep}{\hspace*{5mm} \quad}
\newcommand{\institutesep}{\hspace*{5mm} \quad}

\author{
  Shramay Palta$^{\instA}$$^{\instB}$ \thanks{Work done during an internship at Microsoft Research.} \authorsep Nirupama Chandrasekaran$^{\instB}$ \\ \bf \authorsep Rachel Rudinger$^{\instA}$ \authorsep Scott Counts$^{\instB}$ \\\\
  $^\instA$University of Maryland, College Park \institutesep $^\diamondsuit$Microsoft Research, Redmond\\ 
  \texttt{\{spalta,rudinger\}@cs.umd.edu} \\ \texttt{\{niruc,counts\}@microsoft.com}
}

\begin{document}
\maketitle
\input{sections/00-abstract}
\input{sections/10-introduction}

\input{sections/20-methodology}

\input{sections/30-expertise}
\input{sections/40-impact_experience}
\input{sections/50-related_works}
\input{sections/60-conclusion}
\input{sections/70-limitations}

\bibliography{bib/anthology, bib/custom}

\appendix
\input{sections/appendix}

\end{document}

%% file: style/preamble.tex
\usepackage{framed}
\usepackage{mdwlist}
\usepackage{siunitx}
\usepackage{latexsym}
\usepackage{colortbl}
\usepackage{xcolor}
\usepackage{nicefrac}
\usepackage{booktabs}
\usepackage{fnpct}
\usepackage{amsfonts}
\usepackage[T1]{fontenc}
\usepackage{bold-extra}
\usepackage{amsmath}
\usepackage{amssymb}
\usepackage{bm}
\usepackage{graphicx}
\usepackage{mathtools}
\usepackage{microtype}
\usepackage{multirow}
\usepackage{multicol}
\usepackage{todonotes}
\usepackage{xpatch}
\usepackage{latexsym,comment}
\usepackage[normalem]{ulem}

\newcommand*{\missingreference}{{\Huge \colorbox{red}{?reference?}}}
\newcommand*{\missingcitation}{{\Huge \colorbox{red}{?citation?}}}

\makeatletter
\xpatchcmd{\@setref}{\bfseries}{\missingreference}{}{}
\def\@citex[#1]#2{\leavevmode
    \let\@citea\@empty
    \@cite{\@for\@citeb:=#2\do
        {\@citea\def\@citea{,\penalty\@m\ }%
            \edef\@citeb{\expandafter\@firstofone\@citeb\@empty}%
            \if@filesw\immediate\write\@auxout{\string\citation{\@citeb}}\fi
            \@ifundefined{b@\@citeb}{\hbox{\reset@font\missingcitation}%
                \G@refundefinedtrue
                \@latex@warning
                {Citation `\@citeb' on page \thepage \space undefined}}%
            {\@cite@ofmt{\csname b@\@citeb\endcsname}}}}{#1}}
\makeatother

\newcommand{\gem}[1]{\mbox{\textsc{gem}}}
\newcommand{\abr}[1]{\textsc{#1}}

\newcommand{\hidetext}[1]{}
\newcommand{\ignore}[1]{}

\newcommand{\smallurl}[1]{ \begin{tiny}\url{#1}\end{tiny}}

\definecolor{lightblue}{HTML}{3cc7ea}
\definecolor{CUgold}{HTML}{CFB87C}
\definecolor{grey}{rgb}{0.95,0.95,0.95}
\definecolor{ceil}{rgb}{0.57, 0.63, 0.81}
\definecolor{UMDred}{HTML}{ed1c24}
\definecolor{UMDyellow}{HTML}{ffc20e}


%% file: sections/00-abstract.tex
\begin{abstract}

Using a sample of $25,000$ Bing Copilot conversations, we study how the agent responds to users of varying levels of domain expertise and the resulting impact on user experience along multiple dimensions. Our findings show that across a variety of topical domains, the agent largely responds at proficient or expert levels of expertise (77\% of conversations) which correlates with positive user experience regardless of the user's level of expertise. Misalignment, such that the agent responds at a level of expertise below that of the user, has a negative impact on overall user experience, with the impact more profound for more complex tasks. We also show that users engage more, as measured by the number of words in the conversation, when the agent responds at a level of expertise commensurate with that of the user. Our findings underscore the importance of alignment between user and \abr{AI} when designing human-centered \abr{AI} systems, to ensure satisfactory and productive interactions.
\end{abstract}


%% file: sections/10-introduction.tex
\section{Introduction}


We have seen significant advancements in \abr{LLM} development, which has enhanced the capabilities of model-based agents like ChatGPT that allow them to excel on tasks ranging from quick information retrieval to more creative or technical pursuits such as drafting essays, writing code, and designing artworks. These capabilities are not only attested to by their performance on standardized benchmarks but are also reflected in their use across a diverse set of real-world domains \cite{suri2024usegenerativesearchengines}. Recently developed \abr{LLMs} are able to assist humans across a variety of fields like teaching \cite{wang2024tutorcopilothumanaiapproach, ALSAFARI2024100101} and the clinical domain \cite{han2024ascleai}, and have led to an increase in user productivity \cite{peng2023productivitygithub, cambon2023early}.

\begin{figure}[t!]
    \centering    
    \includegraphics[width=\columnwidth]{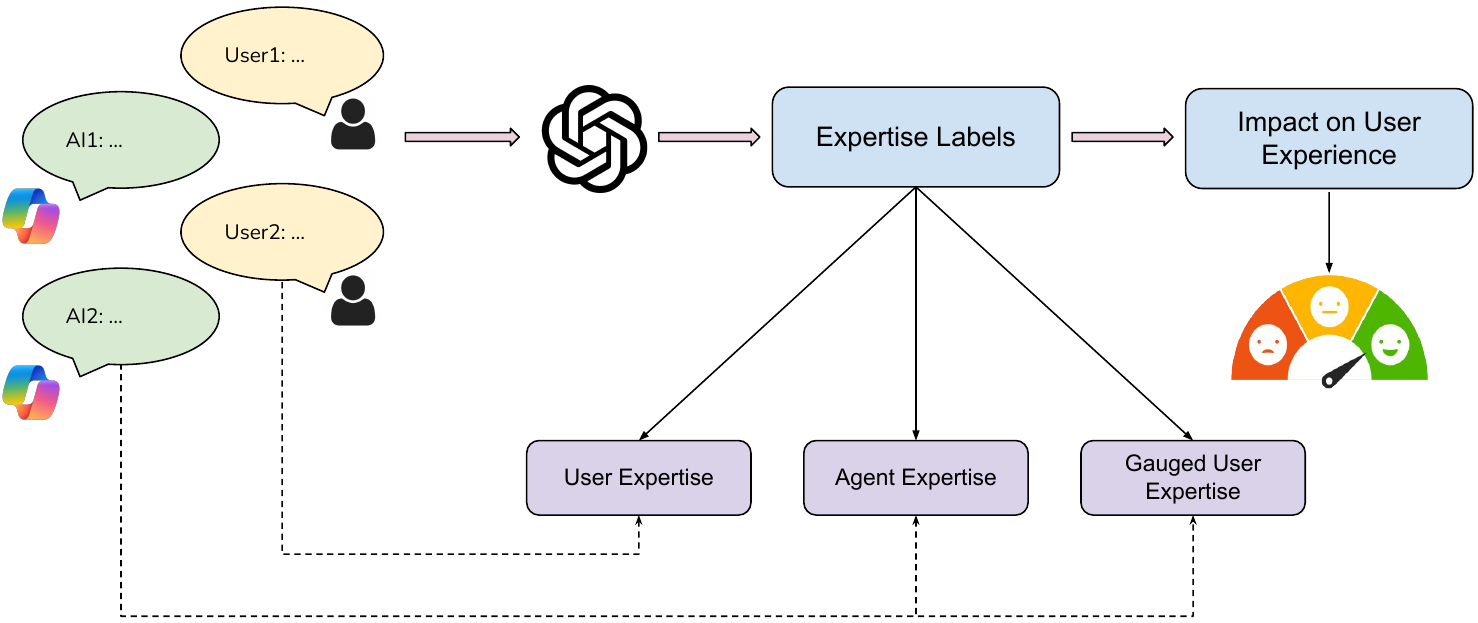}
    \caption{An overview of our expertise classifier pipeline.}
    \label{fig:example}
\end{figure}

However, while millions of people utilize these models for a variety of tasks, their expectations, backgrounds, and interactions with these tools can differ significantly. One key aspect where users might differ is their domain expertise in the conversation topic with the agent \abr{LLM}\footnote{Here on called \abr{LLM}.}. Not all end users share the same level of domain knowledge and thus may have different preferences and abilities to process the information that the model would return to them. A ``beginner'' user might want simple, general purpose information on a topic and could possibly be overwhelmed if presented with too ``high level'' information. On the other hand, a domain ``expert'' might have an unsatisfactory experience with the \abr{LLM} if not given a deeper and detailed response.

Thus, we ask: \textit{What is the ideal expertise level of the LLM, and what are the consequences of any misalignment between the user and the LLM on domain expertise?}

To answer this, we develop an ordinal 5-point scale-based expertise classifier (shown in \autoref{fig:example}) that we apply to a corpus of over $25,000$ Bing Copilot conversations sampled across a variety of domains. We generate three measures of expertise for each conversation. First and second, we classify the level of expertise of the user and the \abr{LLM} respectively in the topic of the conversation. Third, we classify the gauged expertise of the user, defined as the judged level of expertise of the user based on the responses made by the \abr{LLM}. We show similarities and differences in these three types of expertise labels within the same Copilot conversation to identify cases where the \abr{LLM} is aligned or misaligned with the user. We then assess the impact of (mis)alignment on three measures of user interaction experience: user satisfaction, level of engagement, and complexity of task.

%% file: sections/20-methodology.tex
\section{Methodology}

Consider a conversation $C_{i}$ from a corpus of conversations $C$ consisting of $t$ interaction turns of user-agent utterances $C_i = [U_1, A_1, ..., U_t, A_t]$\footnote{Here $U_t$ refers to a \textit{user} utterance and $A_t$ refers to an \textit{AI Agent} utterance.}. We take a random sample of such conversations from Bing Copilot from the month of June 2024 with at least $t \geq 2$ interaction turns for both the user and the agent, yielding a set of $677,801$ conversations. Using the preprocessing steps discussed in Appendix \ref{data_preprocess}, we get a final set of $25,033$ conversations. We present the distribution of the conversations across different domains in \autoref{tab:domain_counts}.


\begin{figure*}[ht!]
    \centering
    \begin{subfigure}[b]{0.49\linewidth}
        \includegraphics[width=\linewidth]{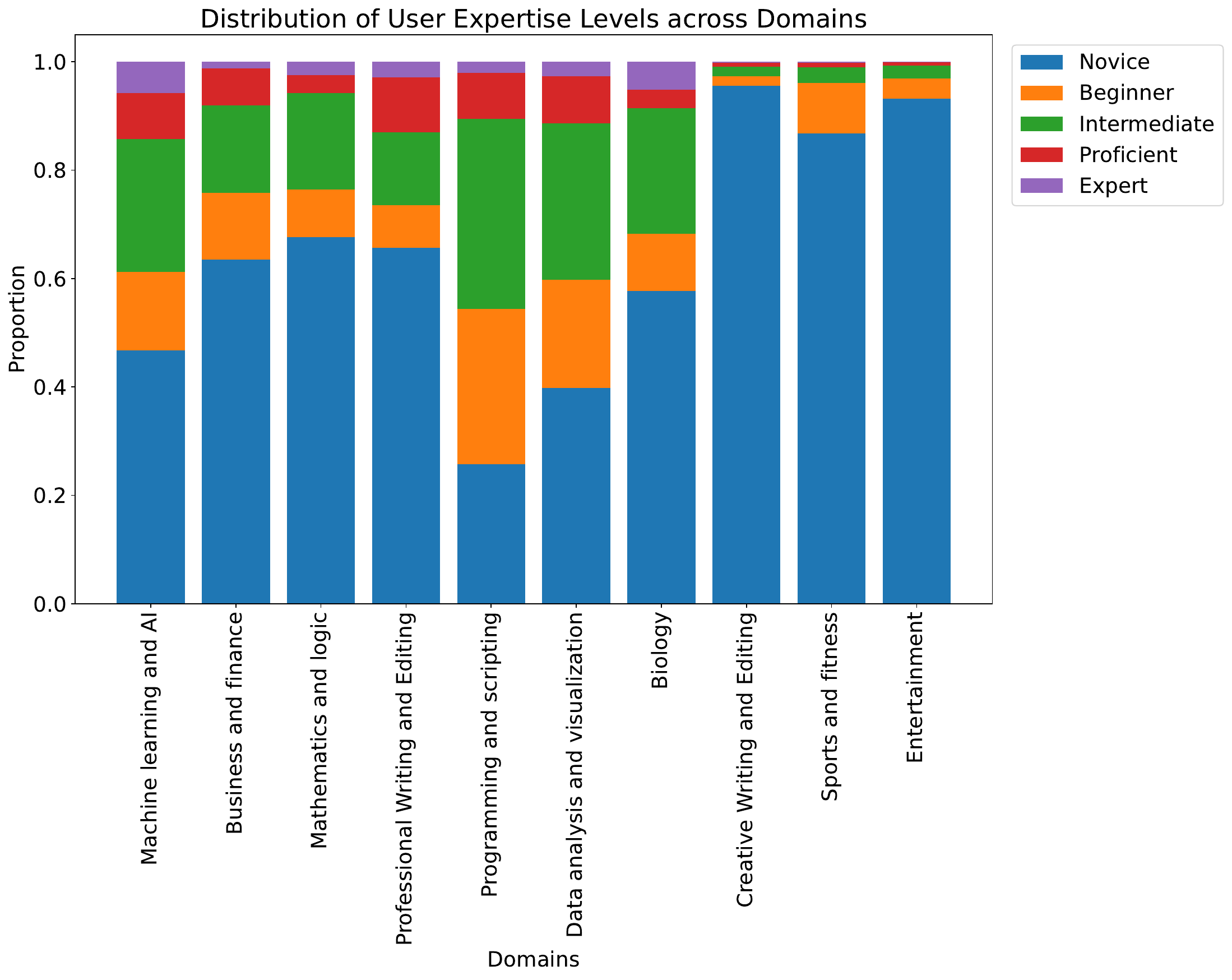}
        \caption{}
        \label{fig:user_dist}
    \end{subfigure}
    \begin{subfigure}[b]{0.49\linewidth}
        \includegraphics[width=\linewidth]{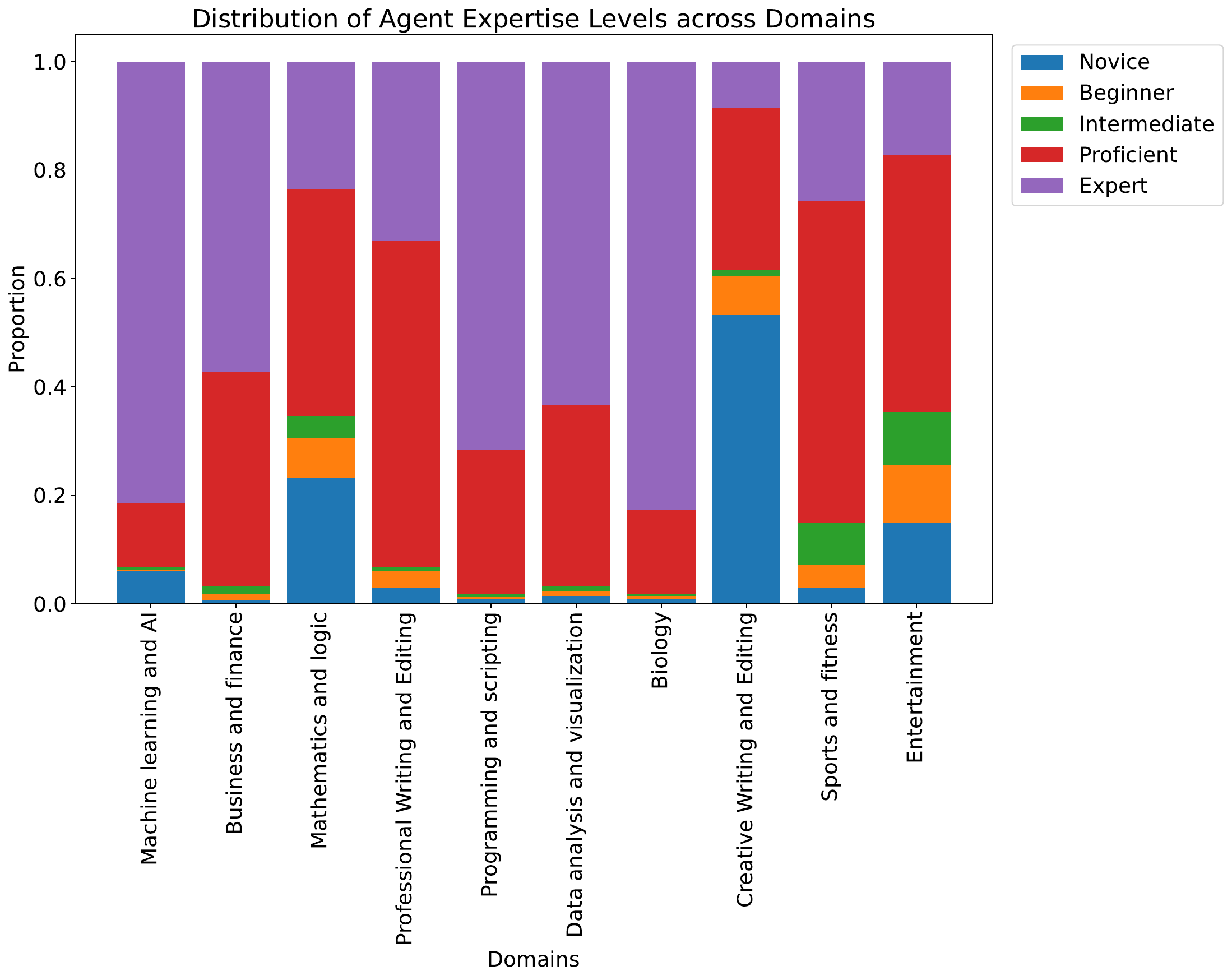}
        \caption{}
        \label{fig:ai_dist}
    \end{subfigure}
    \caption{Barplots showing the distribution of User Expertise (left) and Agent Expertise (right) on different domains of Copilot conversations.}
\end{figure*}

\subsection{Expertise Labels}\label{exp_label}
We expect that users of varying expertise levels interact with Bing Copilot, and thus in order to determine the alignment between the user and the \abr{LLM} expertise, we compute three different types of expertise labels as follows: \\
\textbf{User Expertise}: Expertise of the user in the conversation domain based on the User only side ($U_t$) of a conversation $C_{i}$. \\
\textbf{Gauged User Expertise}: Predicted expertise of the user in the conversation domain, based on the \abr{LLM} ($A_t$) only side of a conversation $C_{i}$. \\
\textbf{Agent Expertise}: Expertise of the agent in the conversation domain, based on the \abr{LLM} ($A_t$) only side of a conversation $C_{i}$.

We prompt \texttt{GPT-4-Turbo} \cite{openai2023gpt4} to compute these three types of expertise labels for each conversation using a $5$-point ordinal scale as ``Novice'', ``Beginner'', ``Intermediate'', ``Proficient'', and ``Expert''. We present the definitions of each of these labels along with the human-validated system prompts in \autoref{fig:user_prompt}, \autoref{fig:ai_prompt} and \autoref{fig:gauged_prompt}. We also provide details on the human validation of the predicted expertise labels in Appendix \ref{prompt_details}.

\subsection{Metrics for User Experience}\label{user_exp_metrics}
We use the following three metrics to understand the impact of expertise (mis)alignment on user experience:
\begin{itemize}[nosep]
    \item SAT Score:~\citet{lin-etal-2024-interpretable} introduced \abr{SPUR} (\textbf{S}upervised \textbf{P}rompting for \textbf{U}ser satisfaction \textbf{R}ubrics), an iterative prompting framework using supervision from labeled examples to estimate the user satisfaction score from a multi-turn conversation with an \abr{LLM} agent. We adopt this framework and define the overall satisfaction score, denoted as \abr{SAT}, as the difference between the satisfaction and dissatisfaction scores\footnote{\abr{SPUR} enables the computation of both satisfaction and dissatisfaction scores.}. The \abr{SAT} score ranges from $-100$ to $100$. The SAT score rubric is human-validated, the details of which are mentioned in ~\citet{lin-etal-2024-interpretable}.
    \item Task Complexity: \citet{suri2024usegenerativesearchengines} introduced a task complexity classification method based on Anderson and Krathwohl’s Taxonomy of learning domains \cite{armstrong2010bloom} which categorizes the task complexity into six levels from lowest complexity to highest: \textit{Remember, Understand, Apply, Analyze, Evaluate}, and \textit{Create}. For simplicity, we group \textit{Remember} and  \textit{Understand} as \texttt{Low Complexity} and \textit{Apply, Analyze, Evaluate} and \textit{Create} as \texttt{High Complexity} tasks. The Task Complexity metric is human-validated, the details of which are mentioned in \citet{suri2024usegenerativesearchengines}. 
    \item Conversation Length: As all our conversations are multi-turn, we look at the number of words across all user turns as a proxy for the user engagement level.
\end{itemize}

%% file: sections/30-expertise.tex
\section{User and \abr{LLM} Expertise}\label{expertise}
Using the labels described in \cref{exp_label}, we compute the user expertise, the agent expertise, and the gauged user expertise on our set of $25,033$ conversations and present the distributions of these expertise labels across different domains in \autoref{fig:user_dist}, \autoref{fig:ai_dist} and \autoref{fig:gauged_user_dist} respectively. 

Overall, we observe that a majority of the users ($63.9\%$) are labeled as ``Novice'' on our ordinal scale, with a small number of users being classified as ``Proficient'' ($5.2\%$) or ``Expert''($1.6\%$). These small numbers of ``Proficient'' and ``Expert'' users occur in the more technical domains (like Programming and scripting) as compared to the non-technical domains (like Entertainment).

We also see that, in a majority of the cases, the \abr{LLM} gets labeled as ``Proficient''($34.9\%$) or ``Expert''($42.4\%$). Once again, we observe the number of ``Proficient'' and ``Expert'' labels to be fewer for non-technical domains as compared to the technical domains. Overall, \autoref{heatmap_size} shows that the user has a lower expertise than the \abr{LLM} in a majority ($80.1\%$) of the conversations. This makes sense in that most users are likely to be non-experts, while the model should have higher expertise in order to provide value to the user.

Finally, for the gauged user expertise (\autoref{fig:gauged_user_dist}), we observe that a majority of the users are labeled as ``Intermediate''($37.2\%$) or above. Comparing this finding with the User Expertise label distribution (\autoref{heatmap_size}) indicates that there are cases of user overestimation, in which user expertise as gauged by the response from the \abr{LLM} is higher than when directly assessing user expertise ($57.4\%$ of conversations), and user underestimation, in which gauged expertise is lower than the user expertise ($4.23\%$ of conversations). We also test our expertise classifier on a sample of WildChat \cite{zhao2024wildchat1mchatgptinteraction} conversations in Appendix \ref{wildchat}, where we observe a similar distribution of labels, hence demonstrating the generalisability of our expertise classifier. 

%% file: sections/40-impact_experience.tex
\section{Impact on User Experience}\label{impact_experience}
\begin{figure*}[ht!]
    \centering
    \begin{subfigure}[b]{0.49\linewidth}
        \includegraphics[width=\linewidth]{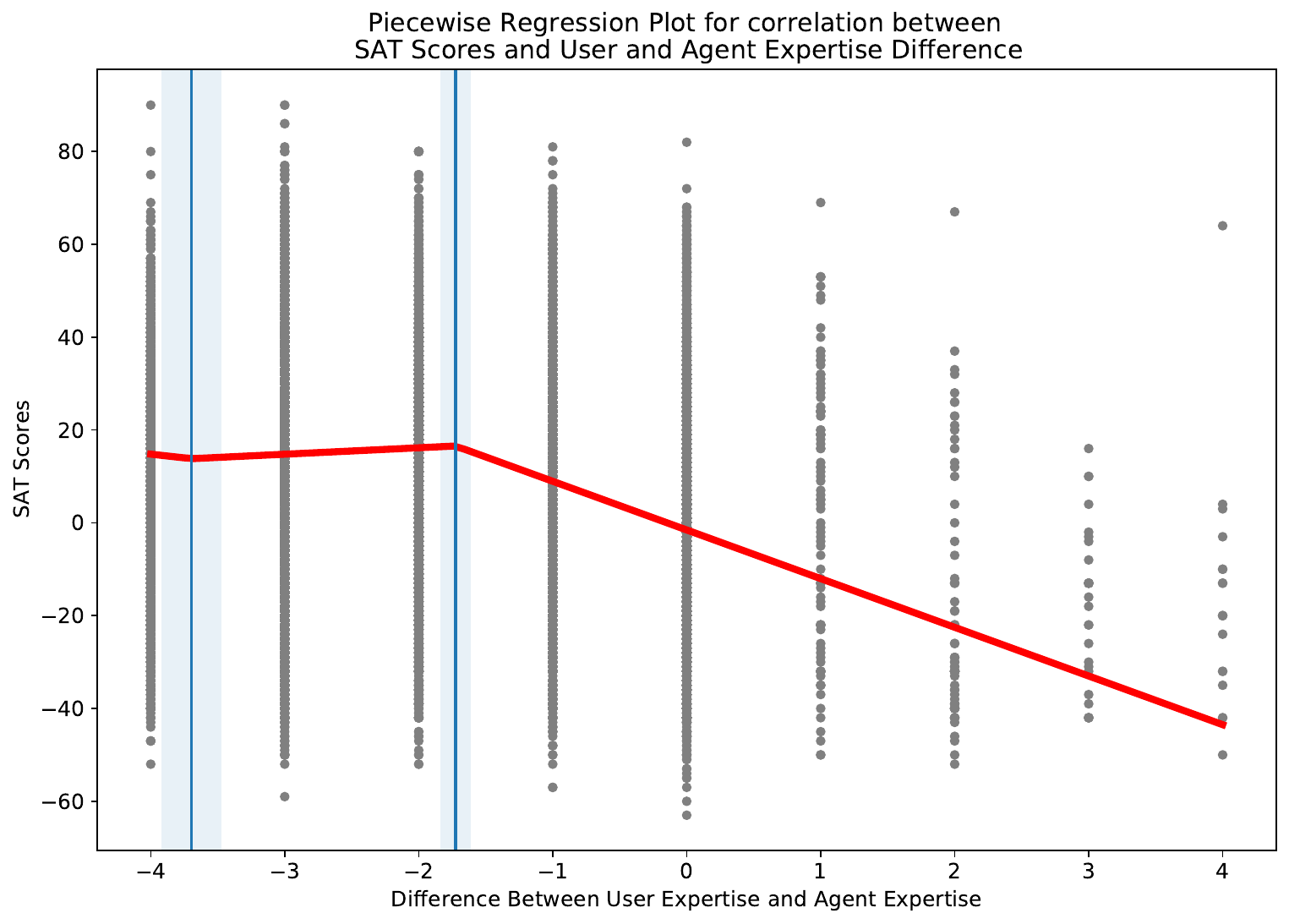}
        \caption{}
        \label{fig:pc_reg_user_ai}
    \end{subfigure}
    \begin{subfigure}[b]{0.49\linewidth}
        \includegraphics[width=\linewidth]{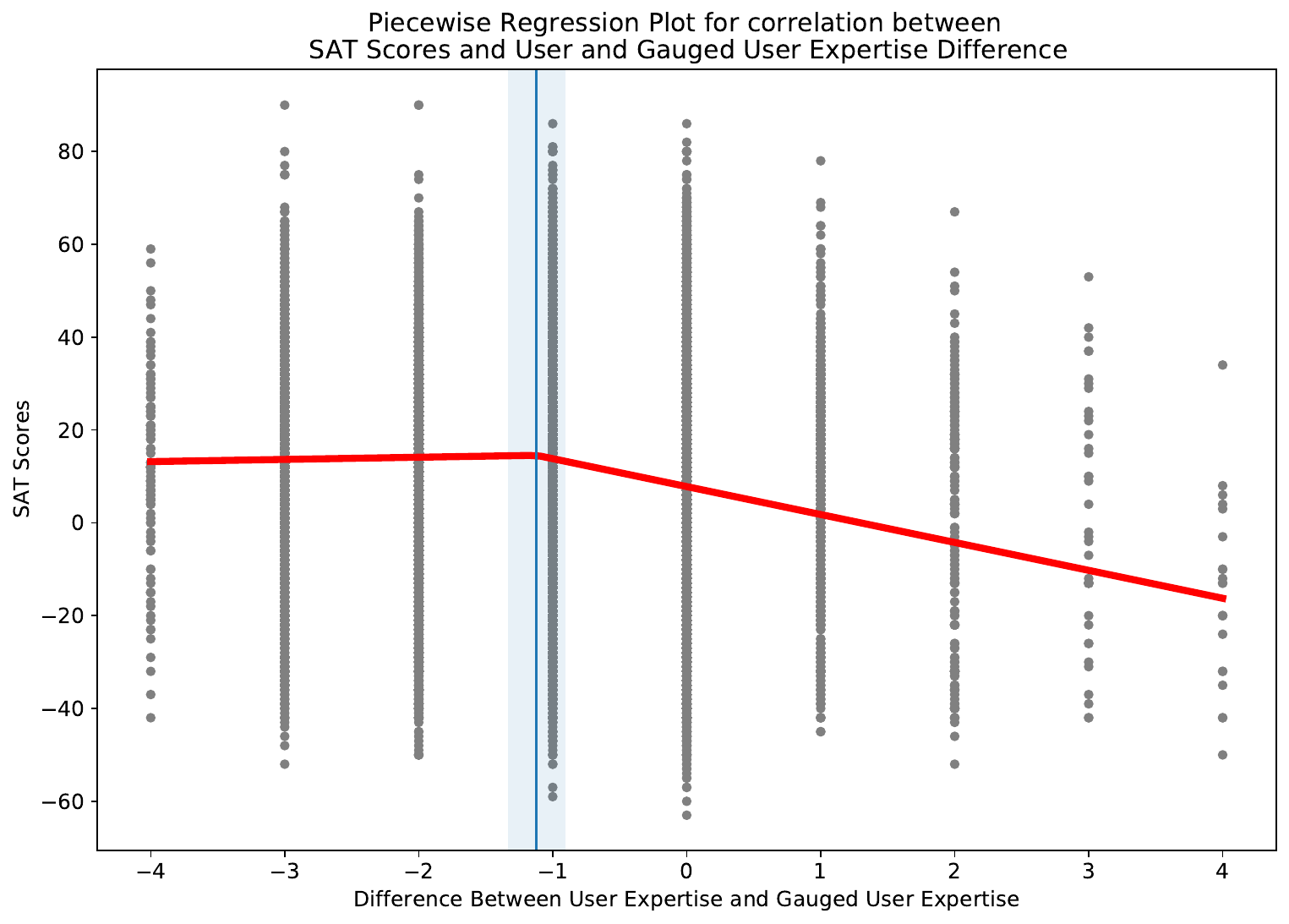}
        \caption{}
        \label{fig:pc_reg_user_gauged}
    \end{subfigure}
    \caption{Piecewise regression plots showing the correlation between Expertise Difference and \abr{SAT} scores.}
\end{figure*}

To understand how expertise (mis)alignment impacts the user experience, we do the following:

First, we fit a piecewise regression of the difference between user and \abr{LLM} expertise on user satisfaction. \autoref{fig:pc_reg_user_ai} shows that when the user expertise exceeds the \abr{LLM} expertise, there is a negative impact $(R^{2}=0.11, p=1.36E^{-141})$ on the \abr{SAT} score, with the overall \abr{SAT} score becoming negative in absolute terms once the user is one level more expert than the \abr{LLM}. Increasing gaps in user expertise over \abr{LLM} expertise are associated with continually lower user satisfaction scores.
~\autoref{heatmap_sat} also highlights that it is preferable for the \abr{LLM} to be quite expert regardless of the user's level of expertise, with lower than ``Proficient'' responses from the \abr{LLM} generally leading to decreased user \abr{SAT} scores. Further, underestimating the user's expertise level (\autoref{fig:pc_reg_user_gauged}), as measured by gauged user expertise, has a clear negative impact $(R^{2}=0.03, p=1.06E^{-26})$ on the \abr{SAT} score, also highlighted in \autoref{heatmap_sat}.

The nature of the user's task may impact the effect of expertise (mis)alignment, and we operationalize one aspect of this with the task complexity measure. We fit linear regressions of the gap in user to \abr{LLM} expertise on user satisfaction separately for conversations with low versus high complexity tasks, with high and low task complexity defined as in \cref{user_exp_metrics}. As seen in \autoref{fig:binary_comp_user_ai}, for low complexity tasks, the user-\abr{LLM} expertise gap had essentially no effect on user satisfaction ($r=-0.04$), while for high complexity tasks, this effect was of medium effect size ($r=-0.29$). The effect of user expertise underestimation as reflected in the gap between actual and gauged user expertise (\autoref{fig:binary_comp_gauged}) shows a similar if somewhat moderated trend with no effect ($r=0.02$) for tasks of low complexity and a small effect for tasks of high complexity ($r=-0.15$) .



Finally, turning to the impact of user-\abr{LLM} (mis)alignment on user engagement, \autoref{heatmap_turn_words} shows that the amount of engagement, as defined in \cref{user_exp_metrics}, increases along with the level of user expertise. That is, more expert users tend to have longer conversations. This effect appears to depend on the level of expertise of the \abr{LLM}, however, such that users at each level of expertise tend to engage more when the \abr{LLM} responds at a similar level of expertise. ``Proficient'' and ``Expert'' users tend to engage relatively more with a Proficient or Expert \abr{LLM}, while ``Novice'' and ``Beginner'' users tend to engage more with a Novice or Beginner \abr{LLM}. Thus while the user satisfaction measure suggests that users prefer a more expert \abr{LLM}, they engage more with \abr{LLMs} of commensurate expertise.

%% file: sections/50-related_works.tex
\section{Related Works}
The term ``expertise'' has been defined along multiple dimensions \cite{bourne2014expertise, garrett2009six}, such as the ``extent and organization of knowledge and special reasoning processes to development and intelligence'' \cite{feltovich1997expertise}. Traditionally, expertise has always been correlated with knowledge, skill and other cognitive concepts \cite{bourne2014expertise}. Building upon this idea, \citet{desmarais1995user} introduced a probabilistic approach to model user expertise, and \citet{dialogue_expertise} did so in a dialogue based setting. Different methods like heuristic rules \cite{vaubel1990inferring} have also shown promise in inferring expertise for word processing tasks. Notably, we model the user, agent and the gauged user expertise from conversational data to show the impacts of expertise mis-alignment on the user interaction experience. 

Additionally, many works have explored how \abr{LLMs} are mis-aligned with humans across an axis of different dimensions like moral judgements \cite{hendrycks2023aligningaisharedhuman, jiang2022machineslearnmoralitydelphi}, cultural and societal norms \cite{palta-rudinger-2023-fork, acquaye-etal-2024-susu, naous2024havingbeerprayermeasuring, bhatia-shwartz-2023-gd, huang-yang-2023-culturally}, healthcare \cite{levy-etal-2024-evaluating} and notions of plausibility \cite{palta-etal-2024-plausibly}. Similarly, we show that \abr{LLMs} are misaligned with humans along notions of expertise, which can lead to unsatisfactory user experiences. 

%% file: sections/60-conclusion.tex
\section{Conclusion}
We examined the alignment between \abr{LLMs} and users along a dimension relevant to the user experience: expertise. We show that the \abr{LLM}'s expertise is largely proficient or expert, which correlated with positive user satisfaction and exceeded user expertise in a majority of the cases. Further, underestimating the user's level of expertise correlated with lower and even negative user satisfaction, with the effect stronger for more complex tasks. Users tended to engage more, however, when the \abr{LLM} responded at a level of expertise similar to their own, suggesting that the system strike a balance between generally high expertise which is liked by all users and matched expertise to best engage users. Future work may explore intervention strategies to strike this balance and mitigate obvious cases of user underestimation in real time.

%% file: sections/70-limitations.tex
\section{Limitations}
Our analysis of user and \abr{LLM} expertise misalignment and its downstream impacts is based on predicted expertise labels and predicted user satisfaction scores (\abr{SAT}). While we human validated the classification labels, there is still a possibility of some errors which could impact the results. 
We study conversations only in English, which may limit the generalizability our findings, given that a lot of conversations with Copilot take place in non-English languages. Our analysis is limited to English conversations to facilitate human-validation of predicted expertise labels. Future work may consider extending our findings in the multi-lingual domain.
We use the same prompts to predict user and agent expertise on all the topical domains of our conversations. While more personalized templates might be able to capture the notions of expertise more accurately for different domains, we use the same template throughout to be able to make a fair comparison across all our experimental settings. 
Additionally, we only use GPT-4 to predict the expertise labels. While it is possible that other \abr{LLMs} might be able to judge the expertise better, we restrict ourselves to the same model as the SAT rubric and the task complexity classifier to maintian model consistency. Future work may consider extending our expertise classifier to evaluate the alignment between humans and multiple different \abr{LLMs}.
Finally, all our results are correlational, but do indicate that a mismatch in expertise between the user and \abr{LLMs} could be one of the causes of user dissatisfaction. We hope our findings motivate future works to involve experiments where agent expertise can be manipulated to determine whether it has a causal impact on user satisfaction or not.

%% file: sections/appendix.tex
\section{Appendix}

\subsection{Data Preprocessing}\label{data_preprocess}
The requirement of two or more turns helped ensure we had sufficient signal to assess the level of expertise of both user and \abr{LLM}. Additionally, the SAT score classifier \cite{lin-etal-2024-interpretable} is also based on multi-turn conversations. We then use the \texttt{langdetect}\footnote{https://pypi.org/project/langdetect/} library \cite{nakatani2010langdetect} to determine the text language at each interaction turn to further filter to conversations where the majority language detected overall was either English, or in case of a tie, English was one of the tied languages. The selection of English only conversations was done to ensure human judges could read a random sub-sample of conversations in order to human-validate our classifications of user and \abr{LLM} expertise.

Using the domain classification methodology introduced in \citet{suri2024usegenerativesearchengines}, we further sampled the conversations at random from a set of topical domains to generate a final set of $25033$ fully anonymized conversations. All personal, private, or sensitive information was scrubbed and masked before the conversations were used for this research. The access to the dataset is strictly limited to the authors who conducted hands-on analysis. This domain filtering step was performed to remove conversations in domains minimally or not at all relevant to the concept of expertise (e.g., travel).

\paragraph{Ethics:}~As part of the production process\footnote{Our use of Copilot conversations falls within the Terms of Use outlined at \url{https://www.bing.com/new/termsofuse}}, the Bing Copilot data is anonymized, and each conversation is formed by aggregating turns based on a unique conversation \abr{ID}. Thus, none of the researchers who analyzed the data are able to recover and identify the conversations from any individual user. In addition, this research study was reviewed and approved by representatives from our institutional review board (\abr{IRB}), as well as our ethics and security teams. No formal \abr{IRB} certificate was required since we did not conduct any human studies for this work.

\subsection{Details on Computational Experiments}
GPT-4 was run on a CPU machine and was allocated three hours to run all experiments with temperature 0. We did not perform a hyper-parameter search. All results are obtained from a single run. For detecting the conversation language, a V100 GPU was used for a total of 1 hour.

\subsection{Usage of AI Assistants}
Our expertise classifier is based on GPT-4. We only use AI Assistants to assist our writing to identify grammar errors, typos and rephrase terms for readability.

\input{tables/domain_counts}

\input{tables/expertise_distributions}

\input{tables/sat_domains}

\subsection{Reproducibility}\label{wildchat}
Owing to privacy concerns, the Copilot conversations used in our study cannot be made public. However, to ensure the generalisability of our expertise classifier, we rerun our expertise classifier on a randomly sampled set of $6400$ conversations from the WildChat dataset \cite{zhao2024wildchat1mchatgptinteraction}. We observe a similar distribution of user, gauged user and LLM expertise labels on this sampled set of conversations. However, it is important to note that, conversations in the WildChat dataset reflect a technological bias, since the service was hosted on Hugging Face Spaces, predominantly attracting developers or individuals connected to the IT domain. As such, it may not fully represent the diversity of conversations that naturally occur with chat-based models. In contrast, Copilot conversations span $25$ topically diverse domains and includes a more representative sample of conversations from the general population.

\begin{figure}[t!]
    \centering
    \includegraphics[width=0.9\columnwidth]{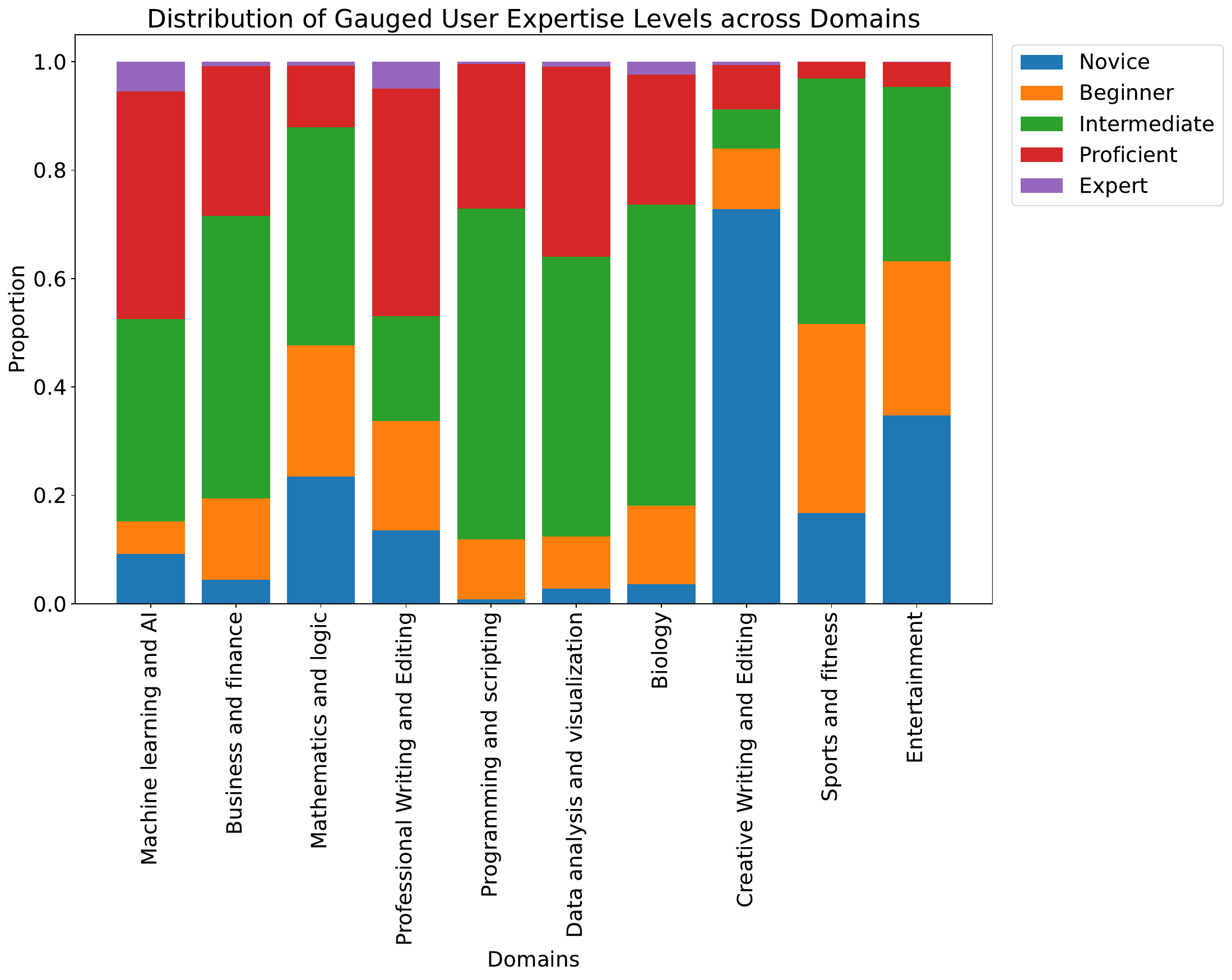}
    \caption{Barplot showing the distribution of Gauged User Expertise on different domains of Copilot conversations.}\label{fig:gauged_user_dist}
\end{figure}

\begin{figure*}[t!]
    \centering
    \includegraphics[width=\linewidth]{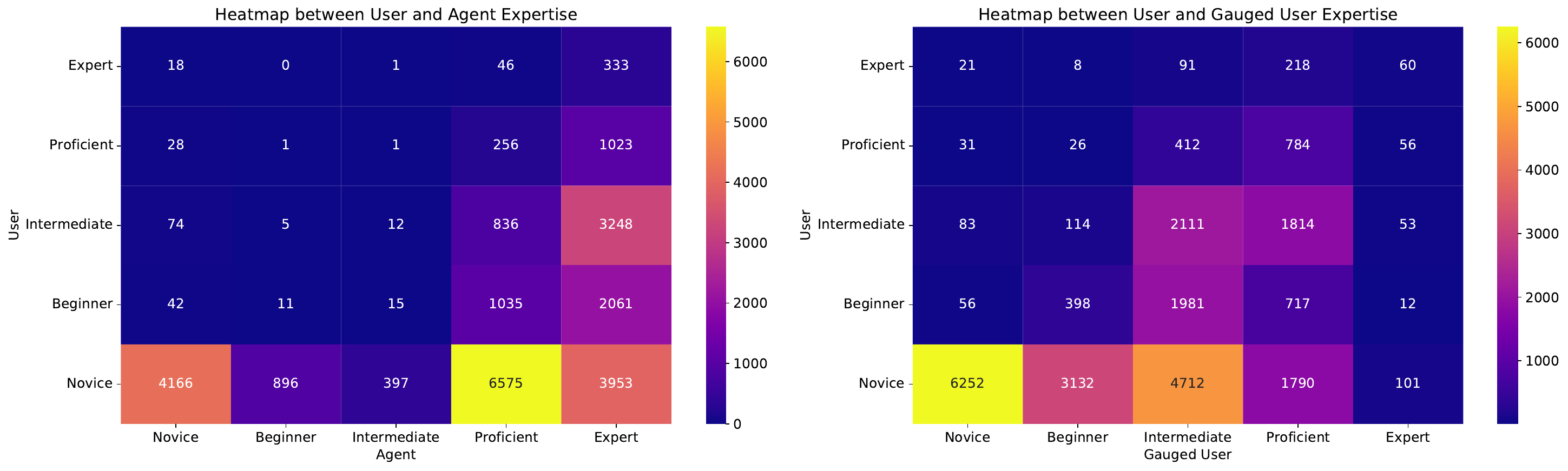}
    \caption{Heatmaps between User and AI expertise (left) and User and Gauged User expertise (right).}\label{heatmap_size}
\end{figure*}

\begin{figure*}[t!]
    \centering
    \includegraphics[width=\linewidth]{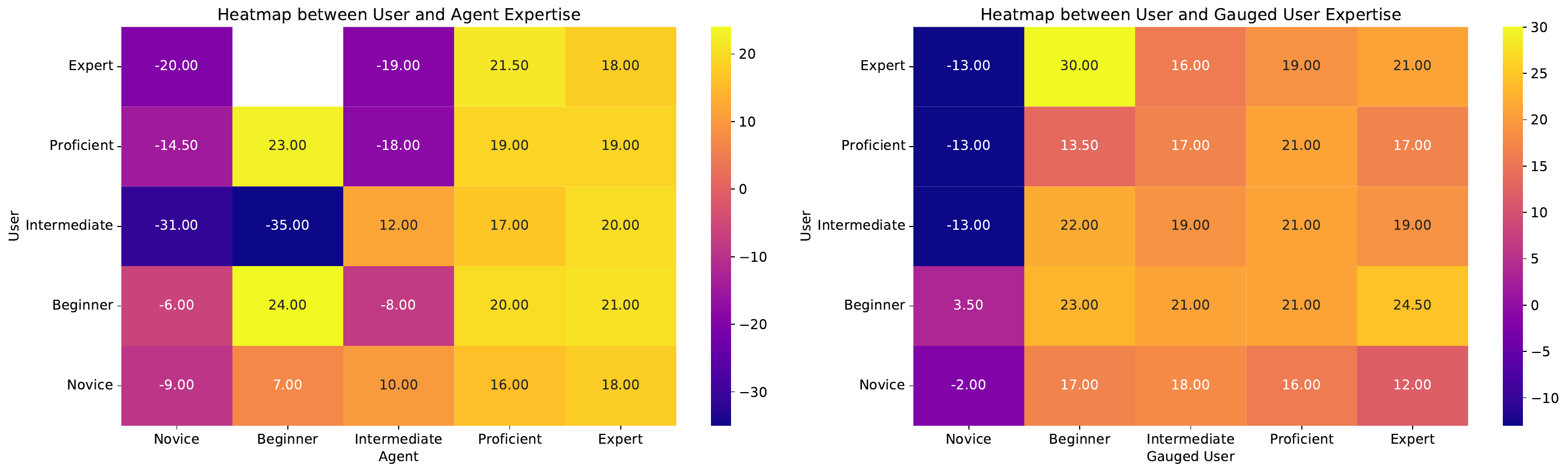}
    \caption{Heatmaps between User and AI expertise (left) and User and Gauged User expertise (right) with density as median \abr{SAT} scores.}\label{heatmap_sat}
\end{figure*}

\begin{figure*}[t!]
    \centering
    \includegraphics[width=\linewidth]{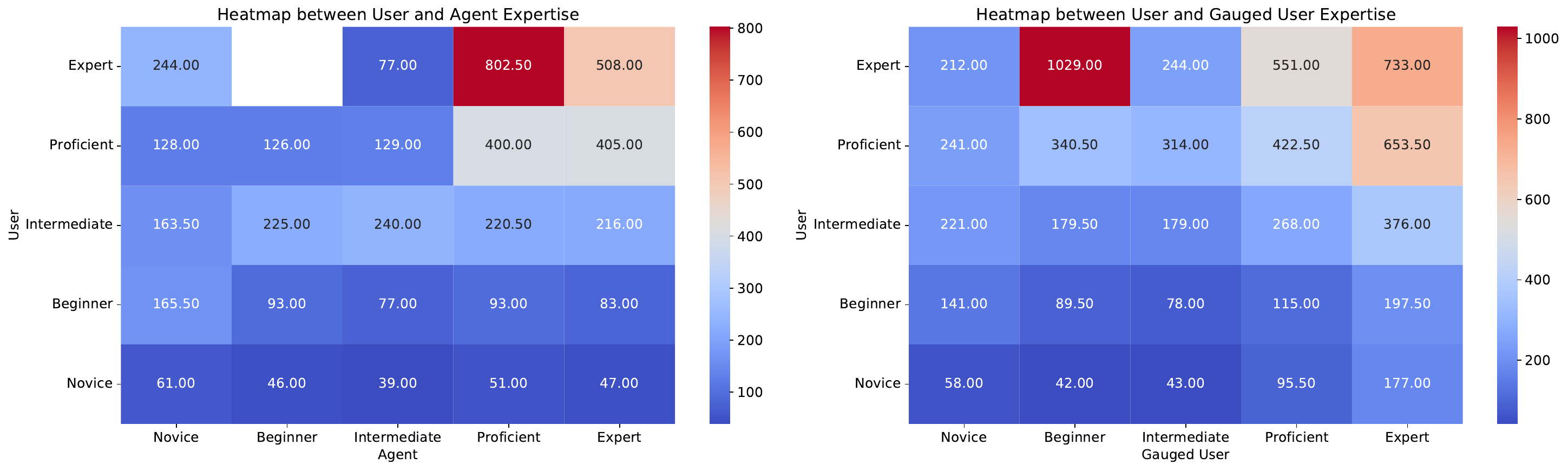}
    \caption{Heatmaps between User and AI expertise (left) and User and Gauged User expertise (right) with density as median number of words in user turns from a conversation.}\label{heatmap_turn_words}
\end{figure*}

\begin{figure*}[t!]
    \centering
    \begin{subfigure}[b]{0.49\linewidth}
        \includegraphics[width=\linewidth]{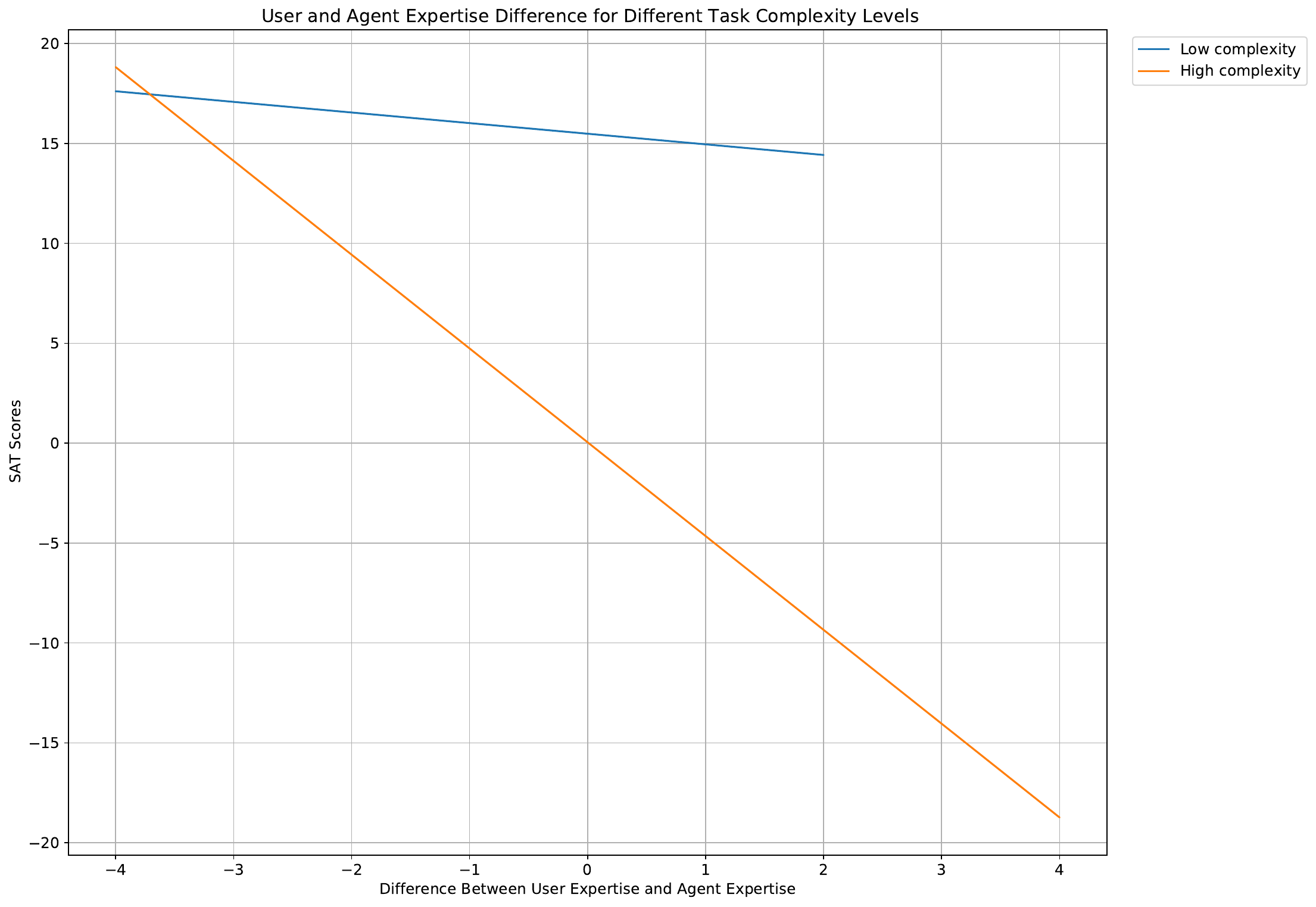}
        \caption{}
        \label{fig:binary_comp_user_ai}
    \end{subfigure}
    \begin{subfigure}[b]{0.49\linewidth}
        \includegraphics[width=\linewidth]{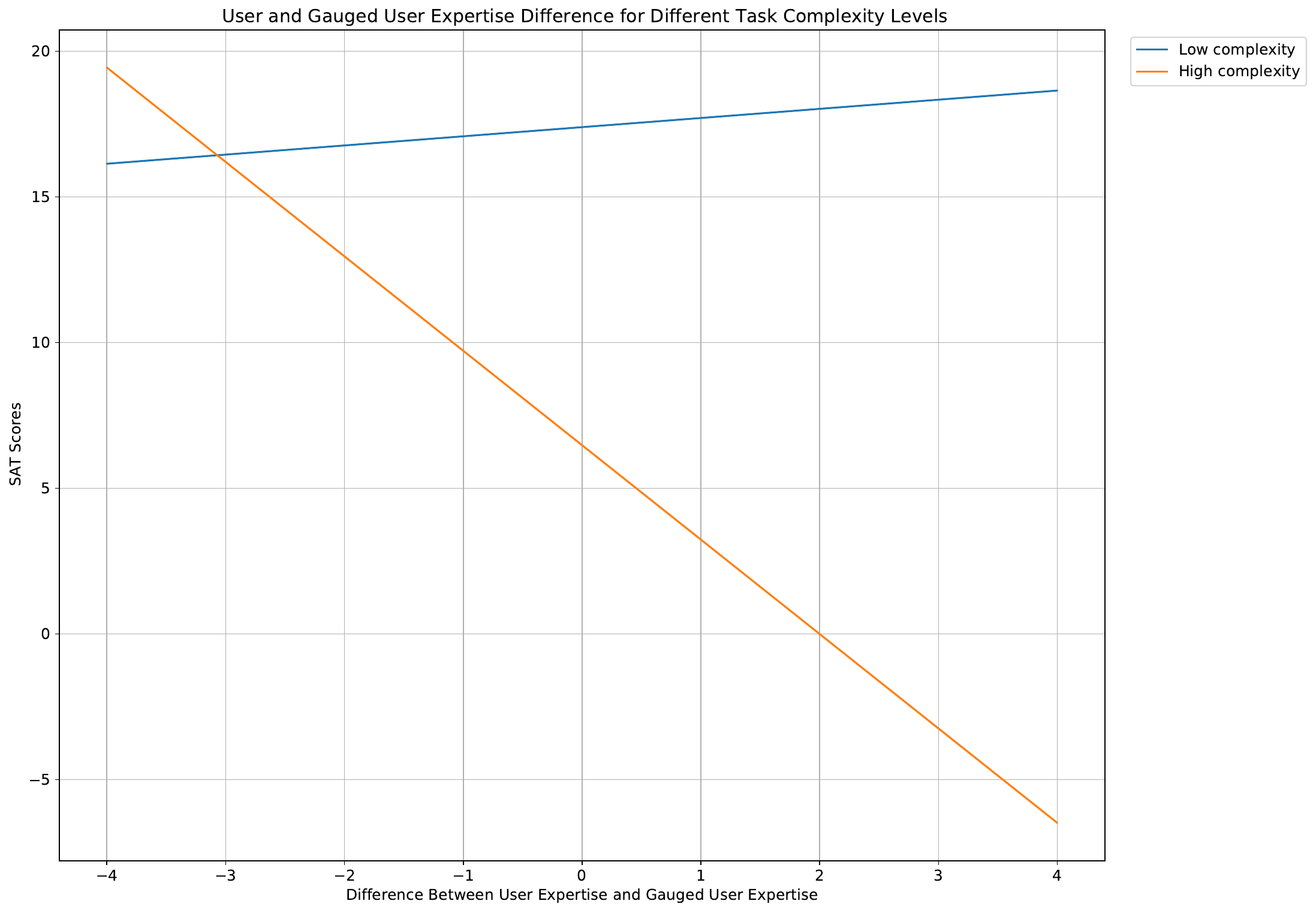}
        \caption{}
        \label{fig:binary_comp_gauged}
    \end{subfigure}
    \caption{Regression plots showing the correlation between Expertise Difference and \abr{SAT} scores for low and high complexity tasks.}
\end{figure*}

\subsection{Prompts}\label{prompt_details}
We release the prompts used for expertise classification. All the prompts were human validated before being used for classification. For human validation of the measured expertise labels, three of the authors independently labeled both the user and \abr{LLM} portions of $50$ Bing Copilot conversations each (total of $300$ assessments: the user and the \abr{LLM} side of $150$ conversations) as "Beginner", "Intermediate", or "Expert". We then compared these labels to a corresponding 3-class version of the final classifier, noting $70\%$ or greater agreement in all cases. The 3-class classifier was used for human validation, as discriminating among the five classes in the full classifier was very challenging for the human labelers.

\clearpage

\begin{figure*}
    \centering
    \fbox{\includegraphics[width=0.75\linewidth]{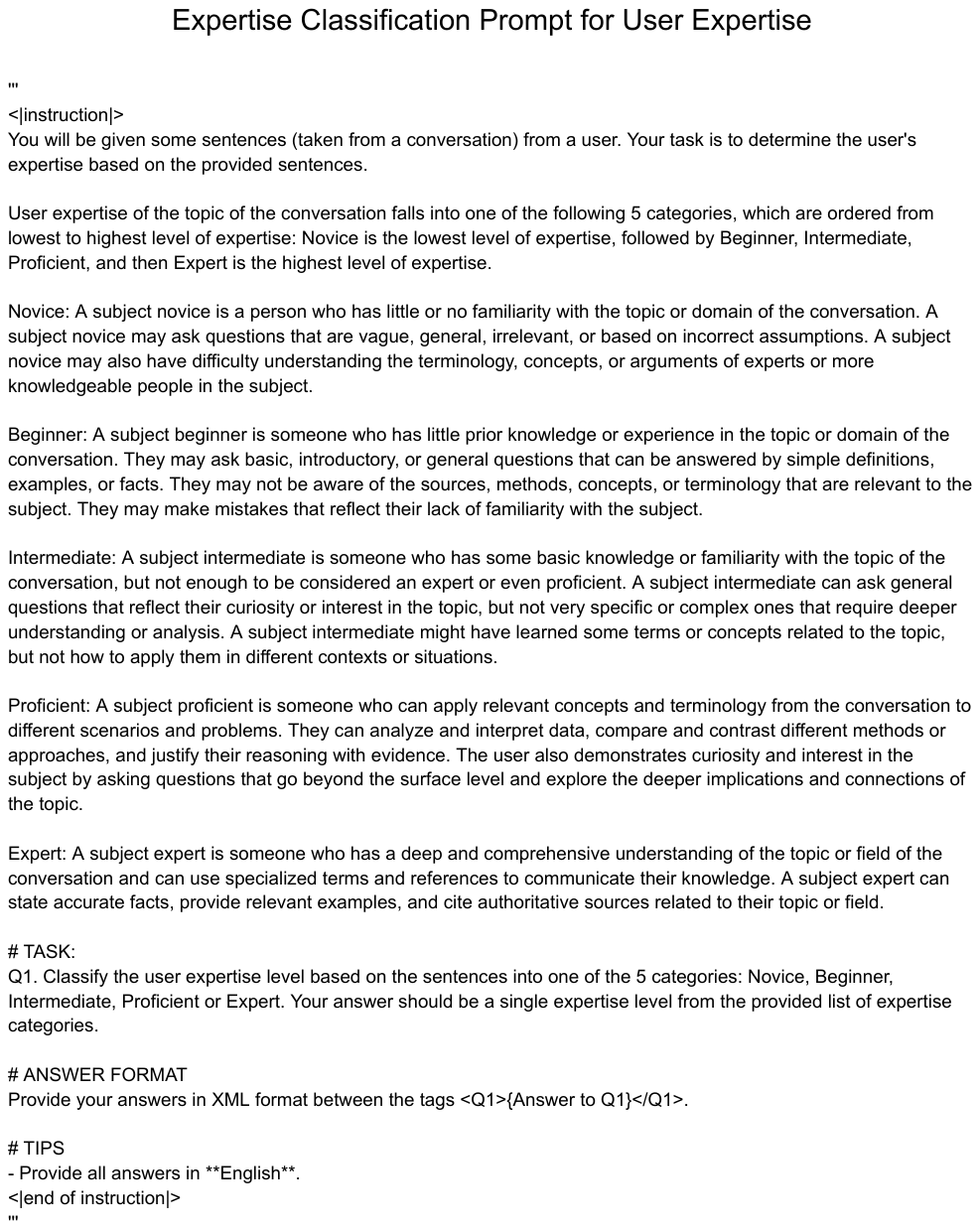}}
    \caption{Prompt for classifying the User Expertise based on the user only turns of Copilot conversations.}\label{fig:user_prompt}
\end{figure*}

\begin{figure*}
    \centering
    \fbox{\includegraphics[width=0.75\linewidth]{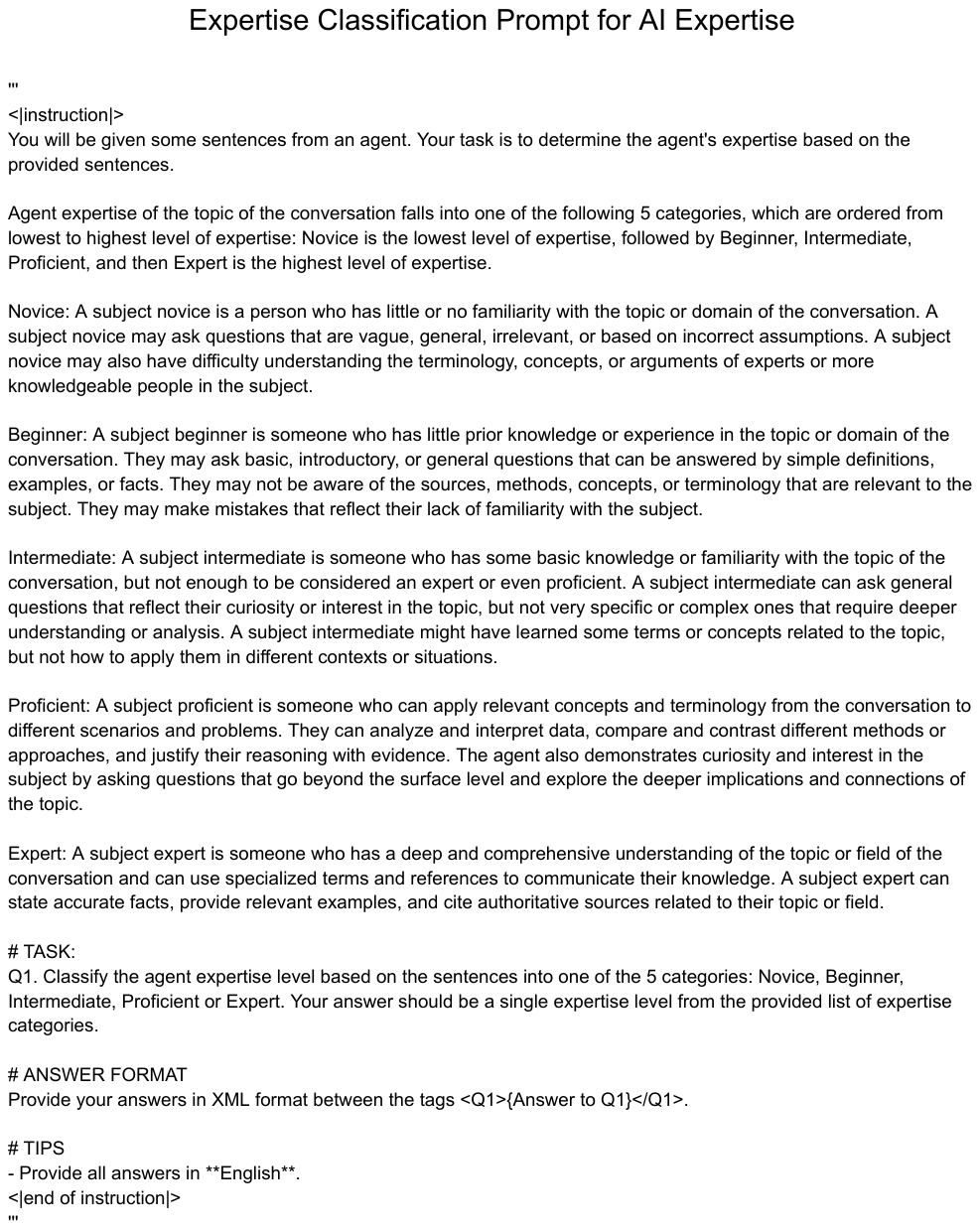}}
    \caption{Prompt for classifying the AI Expertise based on the AI only turns of Copilot conversations.}\label{fig:ai_prompt}
\end{figure*}

\begin{figure*}
    \centering
    \fbox{\includegraphics[width=0.75\linewidth]{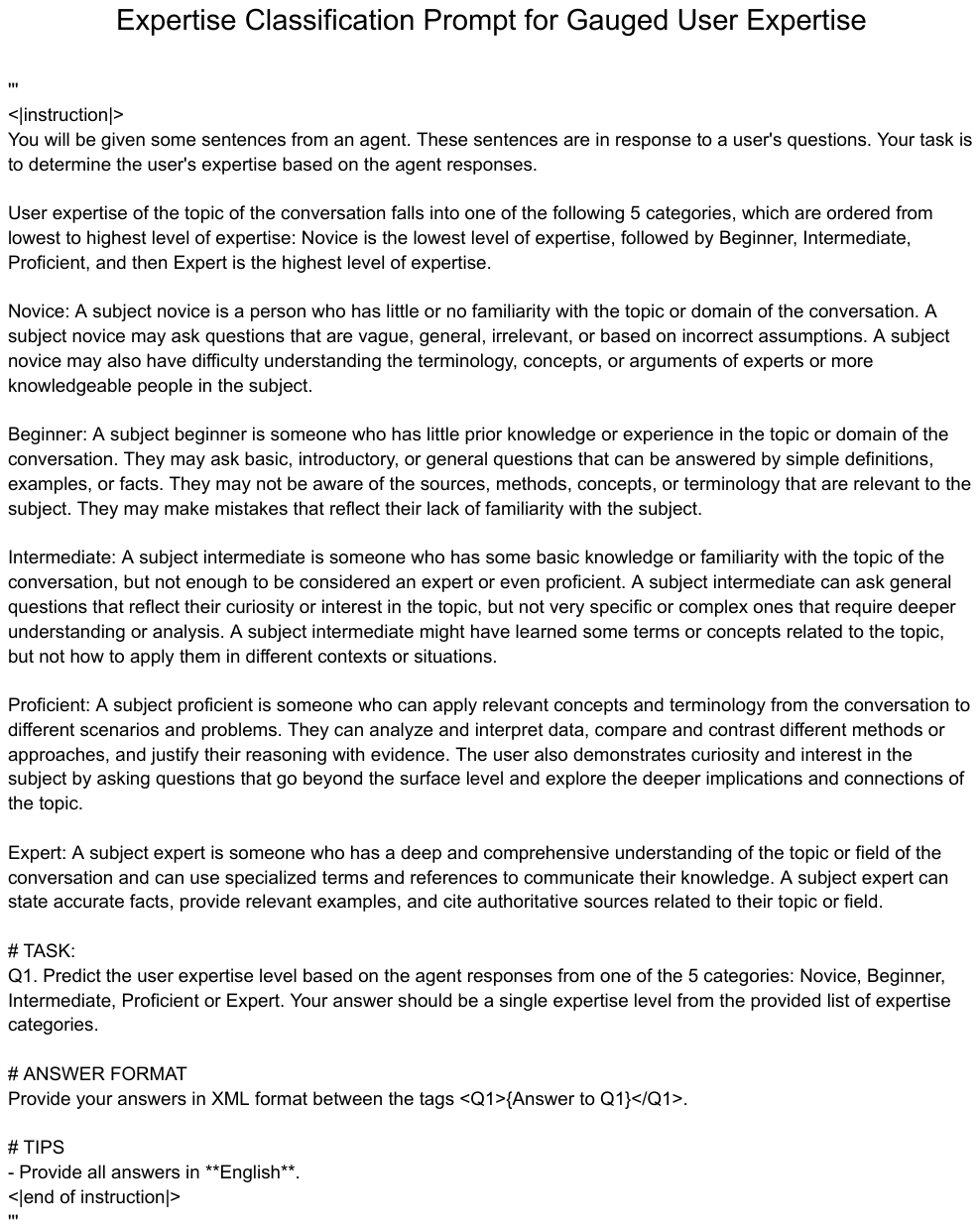}}
    \caption{Prompt for predicting the Gauged User Expertise based on the AI only turns of Copilot conversations.}\label{fig:gauged_prompt}
\end{figure*}

%% file: tables/domain_counts.tex
\begin{table}[ht!]
\centering
\small 
\begin{tabular}{l c}
\toprule
\textbf{Domain} & \textbf{Count} \\
\midrule
Creative Writing and Editing & $6761$ \\
Programming and Scripting & $6078$ \\
Business and Finance & $4090$ \\
Professional Writing and Editing & $2269$ \\
Entertainment & $1406$ \\
Mathematics and Logic & $1306$ \\
Data Analysis and Visualization & $1245$ \\
Biology & $994$ \\
Sports and Fitness & $484$ \\
Machine Learning and AI & $400$ \\
\midrule
\textbf{Total} & $\textbf{25033}$ \\
\bottomrule
\end{tabular}
\caption{Domain-wise distribution of Copilot conversations used in our study.}
\label{tab:domain_counts}
\end{table}

%% file: tables/expertise_distributions.tex
\begin{table}[ht!]
\centering
\small
\begin{tabular}{@{}c|c|c|c|@{}}
\toprule
\multicolumn{1}{c|}{\multirow{2}{*}{\textbf{Label}}} & \multicolumn{3}{c|}{\textbf{Expertise}} \\
\multicolumn{1}{c|}{} & \multicolumn{1}{l|}{User} & \multicolumn{1}{l|}{Gauged User} & \multicolumn{1}{l|}{LLM}\\ \midrule
Novice & $63.9$ & $25.7$ & $17.3$ \\
Beginner & $12.6$ & $14.7$ & $3.7$ \\
Intermediate & $16.7$ & $37.2$ & $1.7$ \\
Proficient & $5.2$ & $21.3$ & $34.9$ \\
Expert & $1.6$ & $1.1$ & $42.4$ \\ \midrule
Total & $100$ & $100$ & $100$ \\ \bottomrule
\end{tabular}
\caption{Percentage distribution of labels for User, Gauged User and LLM expertise on our set of $25033$ Copilot conversations.}\label{expert_label_dist}
\end{table}

%% file: tables/sat_domains.tex
\begin{table}[ht!]
\centering
\small 
\begin{tabular}{l c}
\toprule
\textbf{Domain} & \textbf{Mean SAT Score} \\
\midrule
Creative Writing and Editing & $3.03$ \\
Programming and Scripting & $14.42$ \\
Business and Finance & $17.22$ \\
Professional Writing and Editing & $19.55$ \\
Entertainment & $11.93$ \\
Mathematics and Logic & $-11.03$ \\
Data Analysis and Visualization & $13.35$ \\
Biology & $18.42$ \\
Sports and Fitness & $16.24$ \\
Machine Learning and AI & $16.59$ \\
\bottomrule
\end{tabular}
\caption{Mean SAT Scores for conversations across different domains.}
\label{tab:mean_sat}
\end{table}